\pdfoutput=1
\documentclass[hyphens]{article}
\usepackage{graphicx} 
\usepackage{noto}     
\usepackage{setspace} 
\usepackage{microtype}
\usepackage{doi}
\usepackage[%
]{hyperref} 

\usepackage[font=small, labelfont=bf]{caption} 

\usepackage{cite}

\usepackage{authblk}      

\title{Can Language Models Represent the Past without Anachronism?}

\author[1]{Ted Underwood}
\author[2]{Laura K.\ Nelson}
\author[3]{Matthew Wilkens}
\affil[1]{School of Information Sciences, University of Illinois}
\affil[2]{Department of Sociology, University of British Columbia}
\affil[3]{Department of Information Science, Cornell University}

\date{\normalsize\today}

\usepackage{titling}
\begin{document}

\maketitle

\begin{abstract}
\noindent Before researchers can use language models to simulate the past, they need to understand the risk of anachronism. We find that prompting a contemporary model with examples of period prose does not produce output consistent with period style. Fine-tuning produces results that are stylistically convincing enough to fool an automated judge, but human evaluators can still distinguish fine-tuned model outputs from authentic historical text. We tentatively conclude that pretraining on period prose may be required in order to reliably simulate historical perspectives for social research.
\end{abstract}

\section{Introduction}

Since human history cannot be repeated in a lab, historical research generally works through observation rather than experiment. And observation is enough to illuminate many aspects of the past. To understand why a bill failed, we read documents and count votes. Questions about cultural change, however, may hinge on forms of influence that are hard to observe directly. It is plausible that the New Woman novel changed early twentieth-century attitudes about marriage, but the extent of that influence remains a largely speculative question.

Artificial intelligence could, in principle, cast new light on this kind of question by enabling researchers to simulate the written culture of past eras. If we want to ask how a particular group of writers affected opinions about marriage, we could fine-tune models with and without their works, and measure changes in model behavior. Sensing an opportunity, behavioral scientists have started to try to use LLMs to simulate human populations~\cite{varnum_large_2024, chen_surveying_2024, kozlowski_silico_2024, wang_survey_2024, atari_historical_2023, dillion_can_2023, grossmann_ai_2023, park_generative_2023, xi_rise_2023, lin_agentsims_2023, gao_s3_2023, horton_large_2023, argyle_out_2023, anthis_llm_2025}.

There are reasons to be wary of this approach. Boelaert et al.~\cite{boelaert2024machine} argue that the way LLMs are trained creates inherent ``machine biases'' that misrepresent the diversity of the human population. Similarly, Bisbee et al.~\cite{bisbee2023synthetic} find that LLM-generated outputs exhibit more bias than do actual survey data, particularly underrepresenting minority opinions. 

In this essay we will consider another potential source of distortion, specific to research on the past: the risk that models trained in the present---however carefully prompted or tuned---will import contemporary knowledge, assumptions, styles, or norms into historical contexts where they ought to be absent.

One can, in principle, avoid this risk of anachronism by pretraining models only on material from the past. But pretraining is an expensive, brute-force solution. If historical researchers have to pretrain a cutting-edge model, from scratch, every time they investigate a new social context, LLM-assisted research could demand an impractically large investment of computation and human labor. ``Machine forgetting'' is not likely to be easier to implement---especially not in a context like this where the content being erased is not a single fact but a whole social context \cite{bourtoule_machine_2020, meng_mass-editing_2023, yao_large_2024, hewitt_model_2024, eldan_whos_2023, zhang_negative_2024, pawelczyk_-context_2024, templeton_scaling_2024}. 

If it were possible to get performant contemporary models to simulate the past just through appropriate prompting or tuning, that would be a vastly simpler solution. Our goal in this paper is to provide an initial assessment of the difficulty of that task. Can prompting a model with examples from an older period get it to ``shift temporal gears''? If not, is fine-tuning sufficient?

In describing this as an ``initial assessment of difficulty'' we mean to set a fairly low bar. The paper that follows will not exhaustively test all possible solutions or determine best practices. Rather, it tests a few obvious, simple strategies to see if they work better or worse than expected.

Our conclusions are mixed. 

\begin{itemize}
    \item While in-context learning is generally ineffective,
    \item fine-tuning a model on a few hundred examples can produce a fairly good simulation of period style. 
    \item In head-to-head comparisons, however, human readers still discern the difference between a fine-tuned model and ground truth.
\end{itemize} 
In short, fine-tuning alone is unlikely to erase the risk of anachronism. But it is effective enough to suggest that tuning strategies might eventually be one element of a recipe for producing models that can represent different historical and social vantage points.

\section{Prompting with examples from the period}

Our first experiment tested the limits of prompting, also known as in-context learning. At first glance, language models seem reasonably good at shifting between voices: certainly they can write a Shakespearean sonnet or ``talk like a pirate'' well enough to be amusing. Can they, given examples of prose from an earlier period, continue in the same style?

Human beings are not especially good at precisely dating short passages of text, so for this experiment we relied on a RoBERTa model as a judge. We fine-tuned it to predict publication date using a subset of the Corpus of Historical American English (COHA), 1810--2009.

Then we prompted two models to generate continuations of passages from books that were first published between 1905 and 1914 (inclusive). To test a prompting strategy we used GPT-4o, with a system prompt (illustrated in figure \ref{fig:promptedcontinuation}) that informed the model about the date of the text, as well as a user prompt that gave a substantial paragraph-length excerpt to be continued. We explored both 1-shot and 20-shot prompting.

\begin{figure}[htbp]
  \centering        
    {\fontfamily{phv}
    \fontsize{9}{11}
    \selectfont
    \setlength{\parskip}{\baselineskip}
    \parbox{0.9\linewidth}{\noindent\textbf{System prompt:} Your task is to complete passages from early twentieth-century books. You will be given a passage from a book published in 1913. Continue this passage in the same style, for at least 200 words. Only provide this continuation; do not make any framing remarks, like 'here is the continuation:'.}
    
    \parbox{0.9\linewidth}{\noindent\textbf{Prompt from period text:} In this last case some five or six dollars is economised per minute, for more than twenty yards of film have to be reeled off in order to project during a single minute an object of a person in repose or a landscape. Thus is obtained a practical combination of fixed and moving pictures, which produces most artistic effects. It also enables us to work two cinematographs projecting alternately in order to avoid scintillation, or projecting simultaneously red and green images and reproducing natural colours, thus relieving the human eye, accustomed to receiving the fundamental colours simultaneously, from all physiological fatigue. A word now about the application of cold light to instantaneous photography.}
    
    \parbox{0.9\linewidth}{\noindent\textbf{Continuation by GPT-4o:} The use of cold light in instantaneous photography has revolutionized the approach to capturing images under varying light conditions. Traditionally, photographers struggled with the harshness and variability of natural light or the intrusive nature of flash photography, which could disturb the ambiance of a scene or the natural behavior of subjects, particularly in wildlife and candid human photography. Cold light, with its softer and more diffused nature... allows for a more natural representation of colors and textures, which is often lost under harsh artificial lighting.}
    }
  \caption{Example of a prompted continuation by GPT-4o.}
  \label{fig:promptedcontinuation}
\end{figure}

\subsection{GPT-1914: A pretrained historical language model}

We also generated continuations using GPT-1914, a small (774M) prototype of a GPT-2 model trained only on books published between 1880 and 1914, provided by HathiTrust Research Center. The model was trained on 26.5B tokens using 8 A100 GPUs for 90 hours at an imputed cost of \$1,440, using code by Andrej Karpathy~\cite{karpathy2024llm}. The set of books used to generate prompts was held out from the training corpus, to avoid any risk of memorization. Because this model is quite small, its continuations have less intellectual coherence than those generated by GPT-4o (see example in figure \ref{GPT1914continuation}). However, it proved to be much better than a contemporary language model at reproducing period style.

\begin{figure}[htbp]
  \centering
    {\fontfamily{phv}
    \fontsize{9}{11}
    \selectfont
    \parbox{0.9\linewidth}{\noindent\textbf{Continuation by GPT-1914:} The principle of its action has been already explained (p. 4). We shall here only mention that it can be applied with advantage when we wish to obtain very rapid photographs on paper coated with collodion, as well as those taken by means of gelatine plates. In such cases the exposure must not exceed one second at least; but if the picture is desired to be developed in less time — say half a second — then the temperature should never fall below 20° C., otherwise the image will become too dark after development; moreover, the plate would lose its sensitiveness under these conditions. For ordinary purposes, however, it suffices merely to expose the sensitive surface to a low degree of heat without any special precautions being necessary beyond keeping the}
    }
  \caption{Example of a continuation by GPT-1914, responding to the same passage provided in figure \ref{fig:promptedcontinuation}.}
  \label{GPT1914continuation}

\end{figure}

\subsection{Evaluating in-context learning}

A simple way to observe the difference is to visualize the distribution of perceived ``publication dates'' for ground truth passages, GPT-4o continuations, and GPT-1914 continuations.

\begin{figure}[ht]
  \centering
  \includegraphics[width=\linewidth]{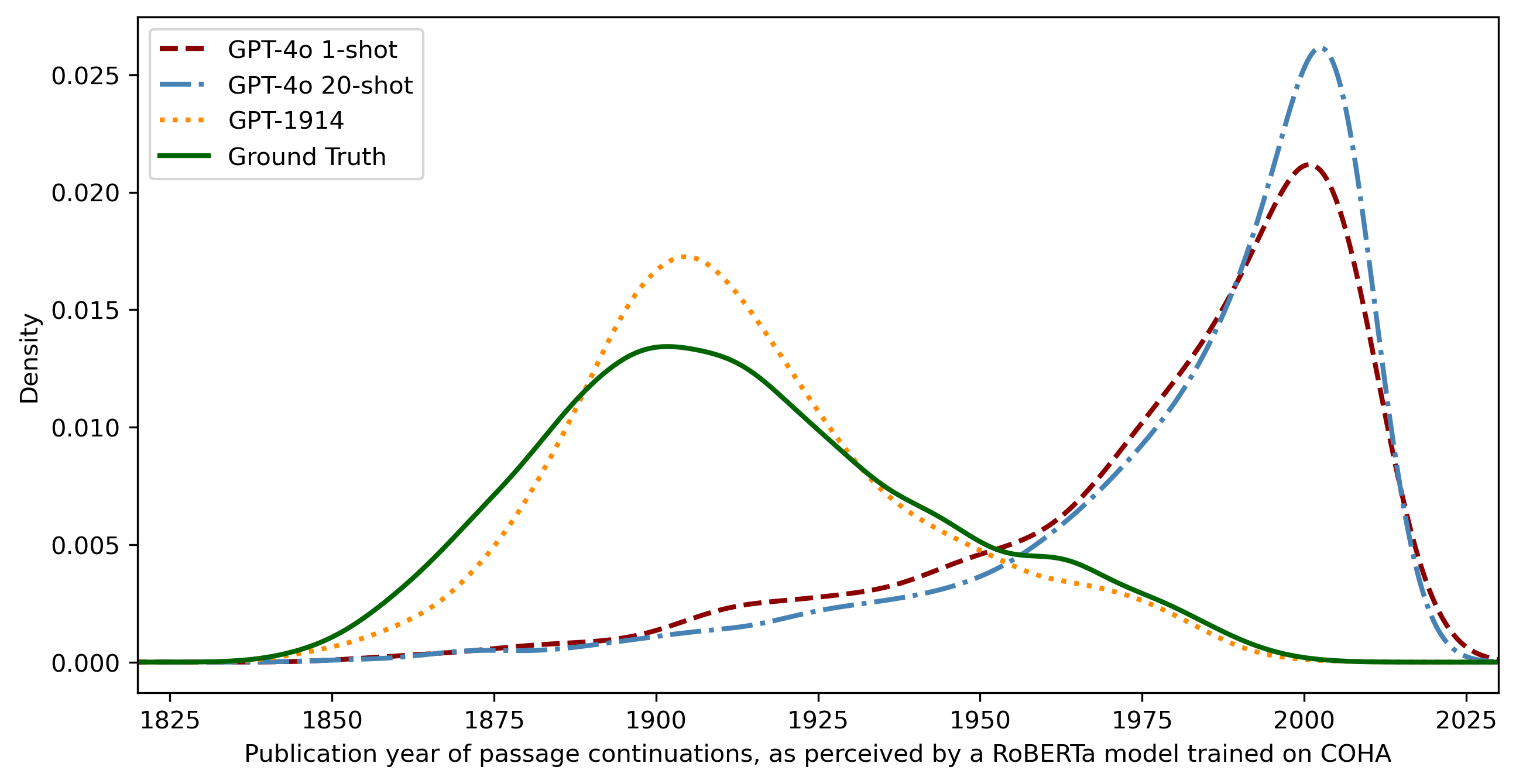}
  \caption{The stylistic center of mass of prompted and period-pretrained models. Kernel density plots for ground truth passages drawn from books, 1905--1914, as well as continuations of those passages by GPT-4o (1-shot and 20-shot) and GPT-1914 (trained only on text 1880--1914).}
  \label{kerneldensity}
\end{figure}

Kernel density plots in Figure \ref{kerneldensity} show the dates estimated by our fine-tuned RoBERTa model. The judging model is not perfect, so distributions spread widely, but ground truth and GPT-1914 both peak in the early twentieth century, while GPT-4o peaks in the twenty-first century (where most of its training data came from). The Jensen-Shannon divergence (JSD) between GPT-1914 and ground truth is 0.006; between GPT-4o (1-shot) and ground truth, 0.310; and between GPT-4o (20-shot) and ground truth, 0.350. Telling the model to emulate examples of an earlier period's style is clearly not effective.

\section{Fine-tuning on examples from the period}

We might expect fine-tuning to be more effective, since it actually changes the model. And if fine-tuning turned out to be an adequate solution, it would certainly be easier than pretraining models like GPT-1914 from scratch. To assess how well this works, we conducted two different experiments. First, we tested fine-tuning using the same judge model employed above to assess the apparent stylistic ``date'' of continuations. Then we ran an experiment with human evaluation.

\subsection{Automatic evaluation by a date-prediction model}

\begin{figure}[ht]
  \centering
  \includegraphics[width=\linewidth]{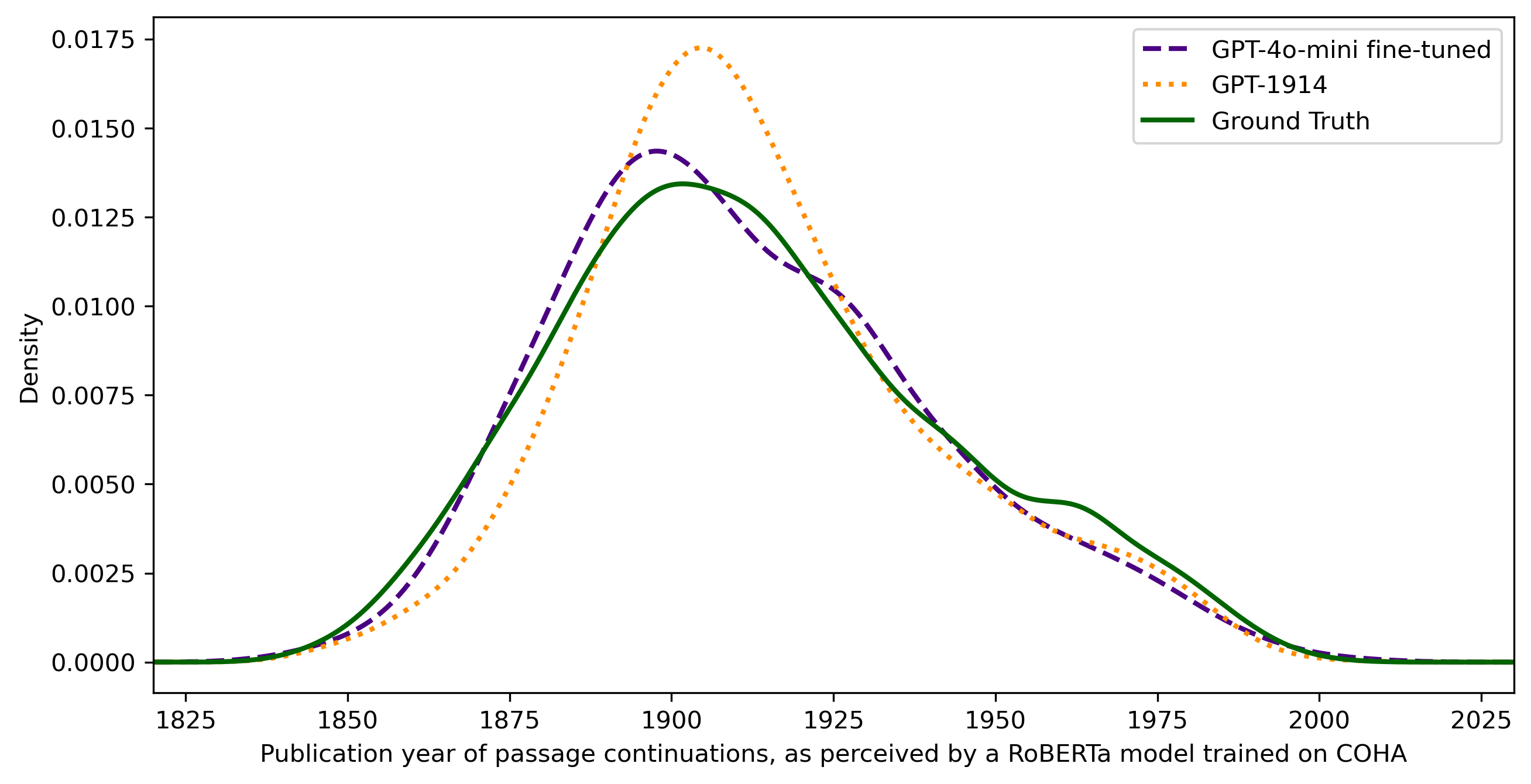}
  \caption{The stylistic center of mass of fine-tuned and period-pretrained models. Kernel density plots for ground truth passages drawn from books 1905--1914, as well as continuations of those passages by GPT-1914 and by a version of GPT-4o-mini fine-tuned on passages 1905--1914.}
  \label{finetunedkernel}
\end{figure}

For a simple test we took roughly two thousand (\emph{passage}, \emph{completion}) pairs from books published 1905--1914 that were not included in our test prompts, and used the OpenAI API to fine-tune GPT-4o-mini on this data\-set. The temporal distribution of completions produced by this fine-tuned model was much closer to ground truth and to a model pretrained purely on late nineteenth- and early twentieth-century texts, as seen in Figure \ref{finetunedkernel}. The JSD between this fine-tuned version of GPT-4o-mini and ground truth was 0.002, very similar to GPT-1914's 0.006.

However, this is not a very sensitive test. The RoBERTa model used as a judge here is only trained to predict a date, not to discriminate authentic passages from anachronistic ones. It probably uses coarse stylistic evidence to make that prediction. Human readers, or larger models, might still be able to detect anachronistic content in passages that superficially sound ``in-period.''

\subsection{Human evaluation}

To frame a sufficiently challenging test, we manually selected 250 passages from early twentieth-century books (1905--1914) addressing topics that are understood quite differently a century later. 

Our list included, for instance, an encyclopedia entry on Alsace (which was then part of Germany) and one on beri-beri (which was then often explained as a fungal disease rather than a nutritional deficiency). While those are differences of fact, we also selected passages that would display subtler differences of attitude, rhetoric, or imagination. For instance, descriptions of non-European places in the early twentieth century tend to slide into racial generalization. A description of sunrise on the moon written in 1913 imagines rich chromatic phenomena, because no one had yet seen photographs of a world without an atmosphere~\cite{moreux1913day}.

We inferred short questions for which these passages might provide an answer, and fine-tuned GPT-4o-mini on the (\emph{question}, \emph{answer}) pairs. (Actually, we fine-tuned five different versions of 4o-mini, holding out a different fifth of the questions in each case.) Then we generated three sets of completions, using the default versions of 4o and 4o-mini, as well as the five fine-tuned models (using each on the set held out from it).

\subsubsection{First evaluation strategy: direct judgment}

Our next task was to define a protocol for comparing these completions to ground truth. Our first attempt at human evaluation asked three expert human coders to examine 120 completions each, and decide in each case whether the completion was plausibly appropriate---or perceptibly anachronistic---for a writer in 1914. This intuitive approach worked poorly, but we feel the failure is illuminating and worth discussing.

After a preliminary round of calibration and discussion, raters agreed about the plausibility of a model continuation 79.4\% of the time. But since verdicts were imbalanced (``plausible" being twice as common as ``not plausible"), this produced a Cohen's kappa score of only 0.554, which might be described as a ``moderate'' level of agreement. Moreover, all the annotators characterized this task as quite difficult, often requiring them to do supplementary research in order to decide whether an answer was consistent with the factual state of the world in 1914. Many questions also required delicate judgment calls about attitude (``How ethnocentric is ethnocentric enough to be plausible for 1914?") One might sum up by saying that readers in 2025 are not especially good at assessing a passage's congruity with the state of the world a hundred-odd years ago. Even for readers with a doctorate, who have studied the period, this is not a sharply marked boundary about which it is easy to reach consensus.

With those caveats, there was a clear ranking of the models in this experiment, as shown in table \ref{tab:model_plausibility}. It is reasonable to infer that fine-tuning helps, but this experiment doesn't allow us to draw very strong conclusions about whether fine-tuning helps \emph{enough}. How good is 0.8? It's difficult to say, in part because there is no baseline here for human error in judging the ground truth itself.

\begin{table}[htbp]
    \centering
    \renewcommand{\arraystretch}{1.3}  
    \begin{tabular}{l r}
        \textbf{model name} & \textbf{plausible} \\
        \hline
        GPT-4o-mini & 0.483 \\
        GPT-4o & 0.642 \\
        GPT-4o-mini fine-tuned & 0.800 \\
    \end{tabular}
    \caption{Direct judgment of model plausibility by annotators}
    \label{tab:model_plausibility}
\end{table}

\subsubsection{Second evaluation strategy: intruder test}

To frame a more interpretable experiment, we constructed an ``intruder test.'' For each question, a new set of expert annotators were given four unlabeled responses (including three model answers along with the ground truth passage), and asked to select the ``intruder''---the passage actually drawn from a period source. This experiment doesn't require annotators to directly judge the plausibility of a model's response. Instead it indirectly illuminates that question by revealing which models succeed in fooling readers.

\begin{table}[htbp]
    \centering
    \renewcommand{\arraystretch}{1.3}  
    \begin{tabular}{l r r}
        \textbf{source} & \textbf{times chosen} & \textbf{fraction chosen} \\
        \hline
        GPT-4o-mini & 6 & 0.04 \\
        GPT-4o & 15 & 0.10 \\
        GPT-4o-mini fine-tuned & 45 & 0.30 \\
        Prose from 1905--1914 & 83 & 0.56 \\
    \end{tabular}
    \caption{How often sources were identified as the original}
    \label{tab:source_selection}
\end{table}

The ranking of models here is the same as in our earlier experiment, but we now have a baseline for the persuasiveness of ground truth, which allows us to see that there is in fact still a fairly big gap, perceptible by humans, between authentic period prose and a model fine-tuned on the period.

A McNemar's test shows that the differences between the bottom three rows of table \ref{tab:source_selection} are statistically significant after Bonferroni correction. Only the difference between the (poorly performing) untuned versions of GPT-4o-mini and GPT-4o fails the significance test.

\section{Conclusions}

We regard the results of our preliminary experiments as both promising and sobering. Disappointingly, we do not find evidence that simply prompting state-of-the-art language models to generate text ``as if'' they were historically situated actors produces convincing results. Prompted generations of this type were judged to be plausible in fewer than two thirds of instances. And that number is likely an overestimate of the models' performance, since it does not penalize cases in which the outputs included telltale disclaimers such as ``in 1914, it is not yet known that \emph{x}'' or ``I am not familiar with \emph{x} as of 1914.'' Such disclaimers occurred in as many as 20\% of generations, depending on the model used.

The poor performance of in-context learning is unfortunate, because these methods are the easiest and cheapest ones for AI-based historical research. We emphasize that we have not explored these approaches exhaustively. It may turn out that in-context learning is adequate---now or in the future---for a subset of research areas. But our initial evidence is not encouraging.

More promisingly, a commercial LLM fine-tuned on roughly 200 in-period passages at a total cost of \$29.50 using API access (that is, without local GPU use) produced outputs that were stylistically similar to our original sources. While fine-tuning requires slightly more work than in-context learning, its demands were far lower than those required to pretrain from scratch even a small and low-performing model like GPT-1914. However, we found that human readers can still distinguish the output of a fine-tuned model from authentic ground truth. Recent work by Fittschen et al. tends to confirm this conclusion, showing that ``leakage'' from pretraining poses real problems for fine-tuned models in historical linguistics~\cite{fittschen_pretraining_2025}. While fine-tuned models are an inadequate solution at present for historical research, they may suffice for some research purposes---for instance, where the goal is simply to avoid stylistic dissonance.

Finally, we observe that GPT-1914, a custom GPT-2--class model pretrained on books published between 1880 and 1914, represents a third path for historical language modeling. Its technical and computational requirements are significantly greater than those of the other two methods employed in our study. (If we had used commercial cloud compute rather than NSF resources, training might have cost \$1,440.) However, this is the only method guaranteed to be free of anachronism. Nevertheless, GPT-1914 in its current form has serious limitations. It is a small fraction of the parameter size of the other models we used, and so falls short of authentic sources in fluency and cogency, even if it remains uncontaminated by later ideas. To improve fluency, we could of course train a larger model at higher cost.

The extent to which verifiably anachronism-free historical models can match the fluency and reasoning ability of models trained on contemporary web-scale data is an open question for future work. In exploring this question, we should also consider the possibility that anachronism may be in some sense unavoidable. Whether we represent the past by instruction-tuning historical models so they can hold conversations, or by teaching contemporary models to ventriloquize an older period, some compromise may be necessary between the goals of authenticity and conversational fluency. There are, after all, no ``authentic'' examples of a conversation between a twenty-first-century questioner and a respondent from 1914. Researchers attempting to create such a conversation will need to reflect on the hermeneutic premise that interpretation always involves a negotiation between present and past, or as Hans-Georg Gadamer puts it, a fusion of two cultural ``horizons''~\cite{gadamertruthandmethod}.

For the time being, we regard our results as consistent with the  hypothesis that simple, low-cost approaches to implementing humanistically credible AI for applied research are not yet sufficient. Further research will be needed before best practices become clear. Part of the challenge may be to construct automated benchmarks, because twenty-first-century human readers (even experts) find it relatively difficult to evaluate the appropriateness of a passage for a given historical period.

\section*{Data availability}

Code and data in support of this article are available at \url{https://github.com/Historical-AI-Lab/GPT-1914}.

\section*{Acknowledgements}

This work was supported by DELTA at NCSA through allocation HUM240002 to TU from the Advanced Cyberinfrastructure Coordination Ecosystem: Services \& Support (ACCESS) program, which is supported by U.S.\ National Science Foundation grants \#2138259, \#2138286, \#2138307, \#2137603, and \#2138296. It was also supported by NEH HAA-290374-23 to MW. The texts of historical books 1880--1914 were provided by HathiTrust Research Center. We are grateful for the expert annotations of model output provided by Christina Han, David Mimno, and Andrew Piper.


\bibliographystyle{arxiv}
\bibliography{main}

\end{document}